\documentclass[conference,onecolumn]{IEEEtran}
\usepackage{cite}
\usepackage[cmex10]{amsmath}
\usepackage{amssymb}
\usepackage{algorithmic}
\usepackage{array}
\usepackage{graphicx}
\usepackage{url}
\usepackage[table]{xcolor}
\usepackage{slashbox}
\usepackage{fixltx2e}
\usepackage{balance}

\newcommand{\eg}{\textit{e.g. }}
\newcommand{\ie}{\textit{i.e. }}

\newcolumntype{L}[1]{>{\raggedright\let\newline\\\arraybackslash\hspace{0pt}}m{#1}}
\newcolumntype{C}[1]{>{\centering\let\newline\\\arraybackslash\hspace{0pt}}m{#1}}
\newcolumntype{R}[1]{>{\raggedleft\let\newline\\\arraybackslash\hspace{0pt}}m{#1}}
\usepackage[caption=false,font=footnotesize]{subfig}

\begin{document}

\title {Technical Report: Implementation and Validation of a Smart Health Application} 
\author{\IEEEauthorblockN{Fran Casino\IEEEauthorrefmark{1}, Constantinos Patsakis\IEEEauthorrefmark{2}, 
Antoni Mart\'inez-Ballest\'e \IEEEauthorrefmark{1}, \\ Frederic Borr\`as\IEEEauthorrefmark{1} and Edgar Batista\IEEEauthorrefmark{1}}
\IEEEauthorblockA{\IEEEauthorrefmark{1}
Smart Health Research Group\\
Dept. of Computer Engineering and Mathematics\\
Universitat Rovira i Virgili\\
Av.Pa\" isos Catalans 26, 43007 Tarragona, Catalonia, Spain}
\IEEEauthorblockA{\IEEEauthorrefmark{2}
Department of Informatics\\
University of Piraeus\\
Athens, Greece \\ \\}
}

\maketitle


\begin{abstract}
In this article, we explain in detail the internal structures and databases of a smart health application. Moreover, we describe how to generate a statistically sound synthetic dataset using real-world medical data.

\end{abstract}

\section{Introduction}
The average age of the world population has experienced a progressive
increase over the last 50 years mainly due to the advances in the fields of
medicine and healthcare. Despite of that, a physically active lifestyle is important in order to reduce/avoid physical issues such as cardiorespiratory problems. The ageing of the society and the need for fostering healthy habits amongst the population imply a great challenge for public healthcare systems. The healthcare sector has always been very active in seeking new technologies that could be integrated so as to improve it. We observe that electronic health (e-health)~\cite{eysenbach2001health} is consolidated and it provides a solid ground for the more recently introduced mobile health (m-health)\cite{Tomlinson2013}. In parallel to this trend, we observe
that the population tends to concentrate in cities and the new concept of
smart cities is gaining popularity at an extremely high rate. From the combination of these two
aforementioned trends a new concept of health (\textit{i.e.} Smart Health) emerged \cite{SmartHealth_Solanas_et_al}. Smart Health is attracting a great deal of attention due to its inherent potential for greater synergy between IoT and healthcare systems. 

\subsection{Contribution and Plan of the Article}

In this article, we provide the details of data structures and databases of our smart health application, which is a route recommendation system. First, in Section \ref{sec:method} we give an overview of the main actors of the system. Next, in Sections \ref{sec:structures} and \ref{sec:data}, we describe the structures and datasets implemented in our system and explain how to generate simulated data using real-world medical statistics, respectively. Finally, Section~\ref{sec:conclusions} concludes the article.

\section{Internal Details of Our System}
\label{sec:method}

In our smart health application, we gather data from different sources such as the citizen's data (\eg their location, route's preferences and medical data), real-time environmental information (\eg air quality, weather information). Moreover, we need to store the route's characteristics in our database. In the next sections, we will describe the internal structures and data management of our application. 

%


\subsection{Data Structures}
\label{sec:structures}
The proposed scheme requires the utilisation of several matrices/tables in
order to store the needed information. 
Table \ref{tab:tableB} is an excerpt of the database that
stores the users' ratings about the utilised real routes.
To compute the top-N route recommendations (\ie a set with the N most promising routes for the citizen), the application applies a Collaborative Filtering technique \cite{Goldberg1992a} based on finding the most similar/closest users. This is
performed using the information of Table \ref{tab:tableB}. Therefore, the system recommends those routes that similar citizens voted with the highest values. 

\begin{table}[!h]
\renewcommand{\arraystretch}{1.4}
\caption{Excerpt of citizens' ratings in $[0,10]$.
The higher the better.}
\label{tab:tableB}
\centering
\begin{tabular}{|>{\bfseries}c| c|c|c|c|}
\cline{2-5}
 \multicolumn{1}{c|}{}& \textbf{$Route_a$} & \textbf{$Route_b$} & $\ldots$ &
 \textbf{$Route_m$}\\
\hline
$u_1$     &2& 4 &$\ldots$ &1  \\
\hline
$\vdots$    & $\vdots$ & $\vdots$ & $\ddots$ &  $\vdots$ \\
\hline
$u_i$     &3& 2 & $\ldots$&  8 \\
\hline
$\vdots$    & $\vdots$ & $\vdots$ & $\ddots$ &$\vdots$   \\
\hline
$u_{n}$    & 6& 5 &$\ldots$ &1 \\
\hline
\end{tabular}
\end{table}
\begin{table}[!h]
\renewcommand{\tabcolsep}{0.25cm}
\renewcommand{\arraystretch}{1.3}
\caption{ Example of table with health information from citizens. Values range from 0 to 1 at intervals of 0.1. Higher values indicate worse health condition. Health issues are only indicative. They are not intended to be a comprehensive list of health issues.}
\label{tab:tableC}
\centering
\begin{tabular}{|>{\bfseries}c| c| c| c|c|c|c|}
\cline{2-6}
  \multicolumn{1}{c|}{}& \textbf{Age} &\textbf{Visual} & \textbf{Respiratory} & \textbf{Reduced} & \textbf{Heart} \\
 \multicolumn{1}{c|}{}& & \textbf{Impairment} & \textbf{Problems} & \textbf{Mobility} & \textbf{Disease} \\
\hline
$u_1$     &23&0.4& 0  &0.8 & 0  \\
\hline
$\vdots$  &$\vdots$   &$\vdots$& $\vdots$  &$\vdots$ & $\vdots$   \\
\hline
$u_i$    &57&0.2&0. 9&0.5& 0.7  \\
\hline
$\vdots$   &$\vdots$ &$\vdots$& $\vdots$ &$\vdots$& $\vdots$ \\
\hline
$u_n$    & 45 &0.2&0.8 &0.1& 0.2  \\
\hline
\end{tabular}
\end{table}
\begin{table}[!h]
\renewcommand{\arraystretch}{1.4}
\caption{Example of real routes data collected in the city of Tarragona. }
\label{tab:tableA}
\centering
\begin{tabular}{L{2cm}C{2.25cm}C{2cm}C{2.25cm}}
\hline
 & $Route_a$ & \ldots & $Route_k$ \\ 
 \hline
{\bf Start Location} & 41$^{\circ}$ 4'44.54"N 1$^{\circ}$12'49.58"E & \ldots & 41$^{\circ}$ 7'45.65"N 1$^{\circ}$14'32.90"E \\
{\bf End Location} & 41$^{\circ}$ 6'32.82"N, 1$^{\circ}$14'58.55"E  & \ldots& 41$^{\circ}$ 8'8.21"N, 1$^{\circ}$14'59.02"E \\
{\bf Distance (km)} & 9.6  & \ldots & 2.32 \\
{\bf Elevation Gain (m)} & 0  & \ldots& 55 \\
{\bf Pavement Quality} & Very good  & \ldots& Average \\
{\bf Status} & Idle  & \ldots& Caution \\ 
\hline
\end{tabular}
\end{table}

\begin{figure}[!h]
\centering
\subfloat[Graphic representation of routes belonging to Tarragona.]{%
\includegraphics[width=0.49\columnwidth]{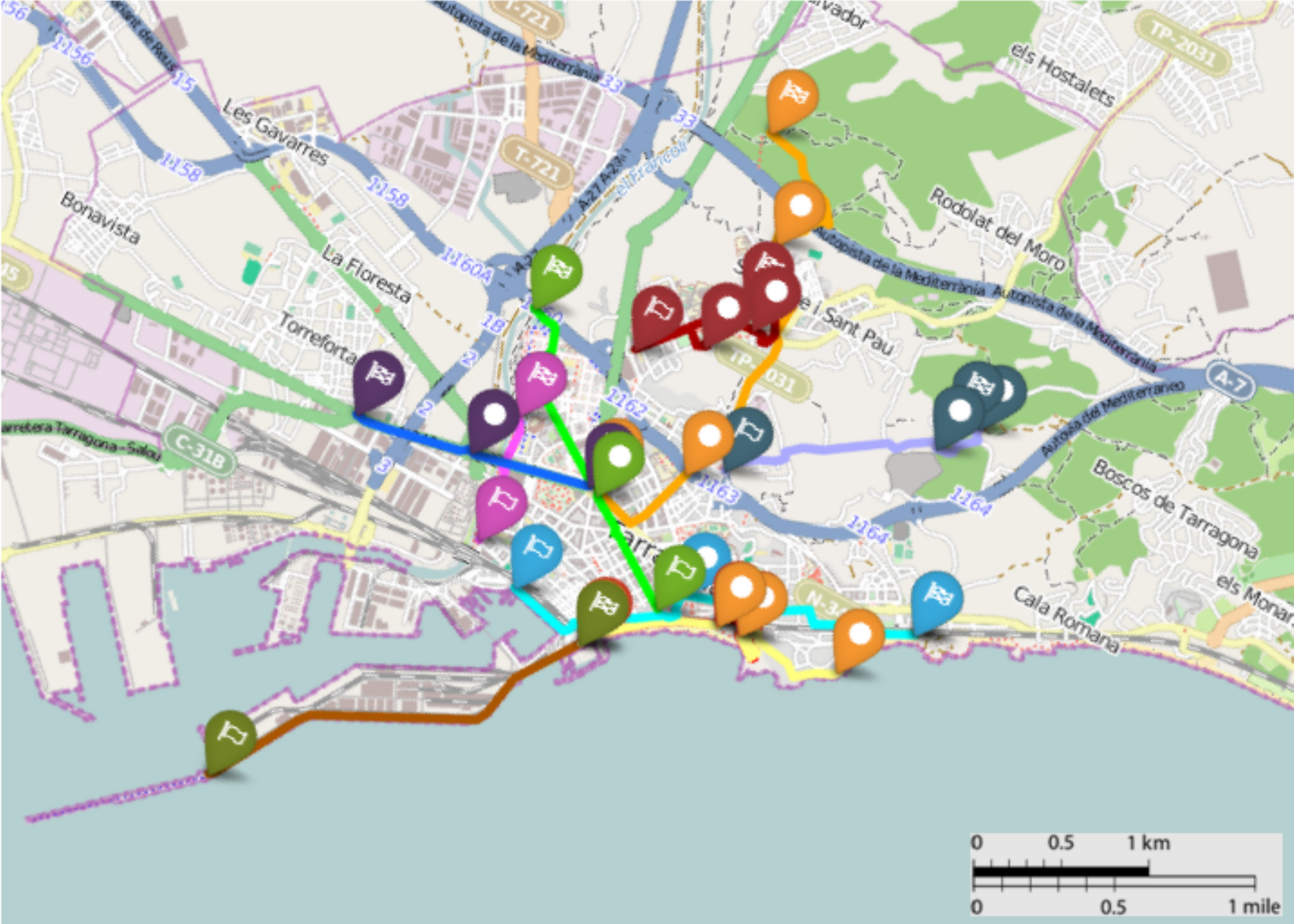}
\label{ruta_tgn}}
\subfloat[Graphic representation of routes belonging to Barcelona.]{%
\includegraphics[width=0.49\columnwidth]{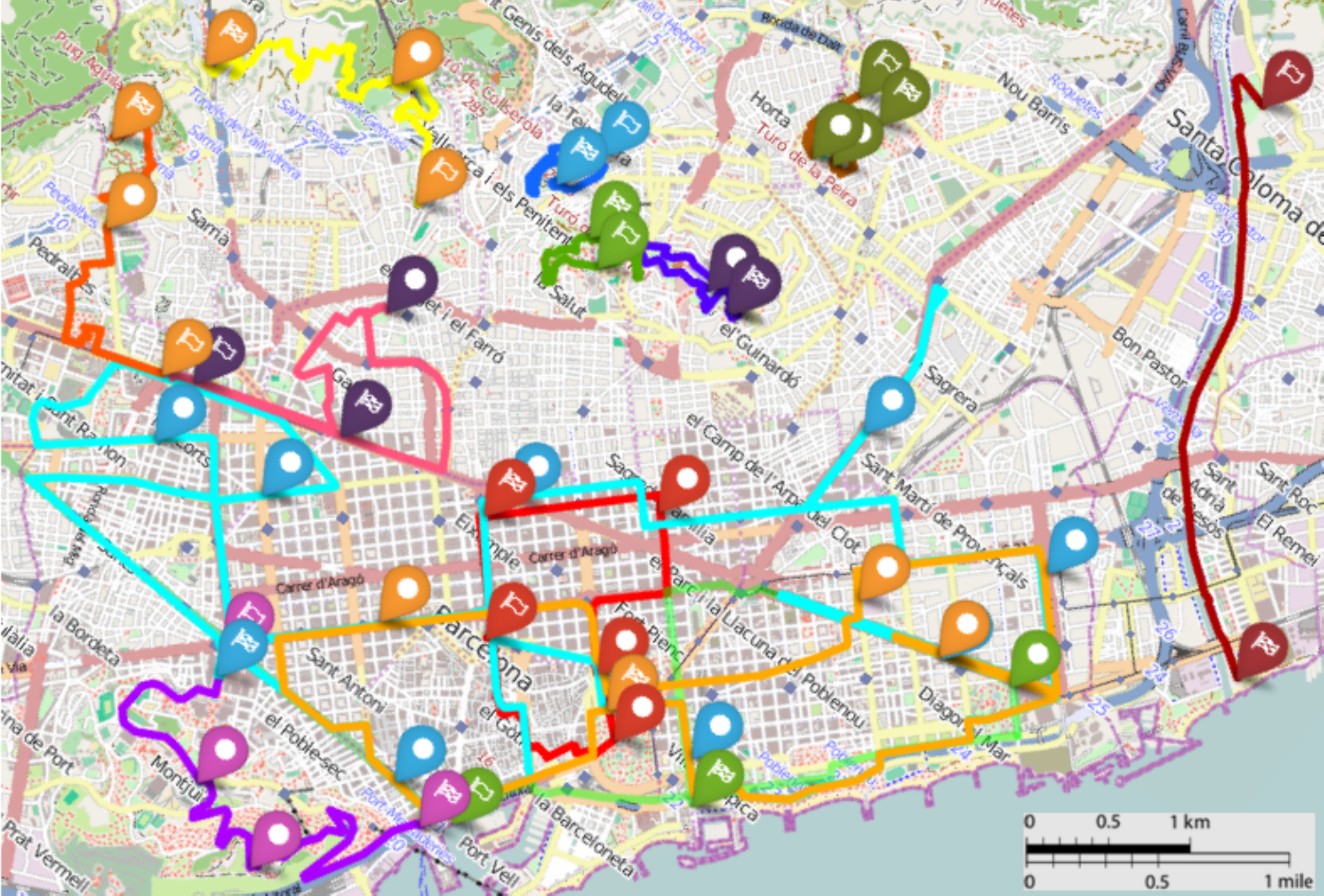}
\label{ruta_bcn}}
\caption{Detailed maps of the routes. Here, each color corresponds to a concrete route. Markers with a blank flag denote the start of a route, whilst markers with a chequered flag indicate the end. Markers with white circles inside correspond to the specific checkpoints of each route.}
\label{rutes}
\end{figure}

Additionally, the system uses a table that contains information about the health condition of
the citizens. Table \ref{tab:tableC} is an example of health information,
which is analysed by the system to avoid inappropriate recommendations.
The information about the routes is also stored. Table \ref{tab:tableA}
is an excerpt of the routes database with information retrieved from a real
scenario in the city of Tarragona. A representation of routes in the city of
Barcelona and in the city of Tarragona are depicted in Figure~\ref{rutes}.
The routes contain checkpoints (\emph{i.e} intermediate route points) which add dynamism to routes and clarify possible loops.

\section{Data Collection and Simulation} 
\label{sec:data}
First, we simulated some data to fill two
databases of preferences and health conditions. In what follows, we describe the data simulation procedure. 

In order to test our proposal we created two databases that store the
preferences of simulated users in the cities of Tarragona and Barcelona. Tarragona database stores the
preferences of 1,000 simulated users about 11 real routes, represented in Table~\ref{tab:tableA} and depicted in
Figure~\ref{ruta_tgn}. In the case of Barcelona, the database stores the preferences of 50,000 users over 28 designed routes, graphically represented in Figure~\ref{ruta_bcn}. Therefore, Barcelona dataset stores a total of 1,400,000 ratings and, hence, is two orders of magnitude larger than Tarragona dataset. It is worth to emphasise that since the stored routes contain checkpoints, the real amount of routes and sub-routes is much higher. 
The simulated data have been generated as follows:

\noindent \textbf{Citizens simulation}: In order to simulate the citizen's profiles, we select their age between 18 and 90 by using a distribution according with the age pyramid of the country \cite{cia}. Also, we consider four main health issues, namely
visual impairments, respiratory problems, reduced mobility and heart diseases.
Clearly, these health issues do not cover all the possible illnesses that
a citizen might suffer, they are only indicative and are not intended for a
precise characterisation of citizens. On the contrary, their goal is to illustrate the operation of our approach in a realistic and practical scenario.
In a further embodiment of our approach, more health problems could be added.
However, for the sake of clarity we have kept this example simple.
To decide whether each citizen suffers from a given disease, we simulate it by using real data provided by the World Health
Organisation~\cite{who} and the World Heart Federation\footnote{\url{www.world-heart-federation.org}}.
Specifically, it we consider that the probability of suffering from
visual impairments is 3.4\%, the probability of having respiratory problems
is 3.2\%, the probability of suffering from reduced/limited mobility is 2\%,
and the probability of suffering from some kind of cardiovascular disease is 14\%. In
the case of heart diseases, we consider that mainly people over $45$
years are affected, since the probability of finding this kind of issues in
younger people decreases drastically and is mainly negligible.

\begin{figure}[!h]
	\begin{center}
		\includegraphics[width=0.5\columnwidth]{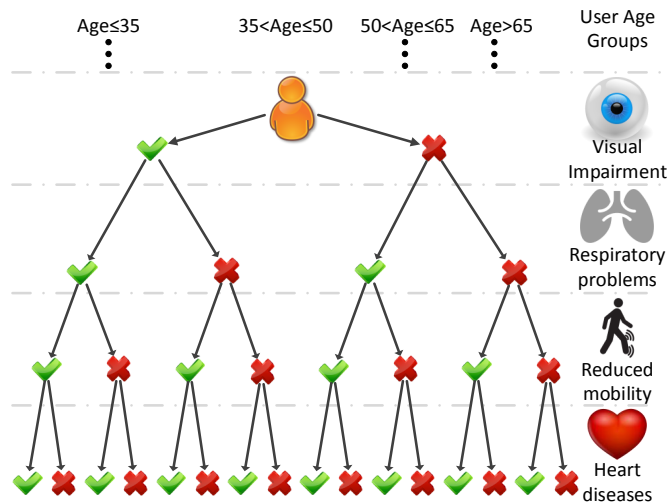}
		\caption{Binary classification of the possible citizens profiles. We
		consider 4 age intervals. For each age interval we distinguish 16
		profiles that represent the possible combinations of suffering
		a given disease.}
		\label{bin_tree}
	\end{center}
\end{figure}

\noindent\textbf{Profile characterisation}: After being simulated, citizens may be
classified in different profiles depending on their health issues and age. A
binary tree representing this classification is depicted in
Figure~\ref{bin_tree}. It can be observed that there exist 16 categories and 4 age
intervals. This results in 64 different profiles.

\noindent\textbf{Ratings simulation}: It is assumed that citizens belonging to a
given profile have similar needs and, thus, would have similar ratings.
Based on this, a relationship between the routes and
the citizens profiles can be established.
$ $\\
\noindent \textbf{a) Determine the range of routes' features}:
The values of the features of each route (\textit{i.e.} distance,
elevation gain and pavement quality) are classified in a range between 0
and 5. In this case, the higher the easier (better) for the citizen.
For instance, if the elevation gain is 0 meters, this feature of the
route will be classified as 5 (very easy). On the contrary, the highest
elevation gain will be classified as 0 (very difficult). The equivalences between features values and ranges have been normalised to fit in the aforementioned range. Note that features are considered independently.

\noindent  \textbf{b) Assigning citizens skills}: Depending on the citizens age
and their health conditions, they will react differently to the routes
features. For example, citizens suffering from heart diseases would be quite
reluctant to follow routes with high elevation gains. To quantify this for example, we assign two negative skill points (related to the elevation gain)
to every citizen suffering from heart diseases at the highest level (\textit{i.e.} 1, according to Table \ref{tab:tableC}). Therefore, the value of each health condition is multiplied by the values of the skills assignment heuristic represented in Table~\ref{tab:tableSkills}.

\noindent  \textbf{c) Determine the ratings}: Finally, to create
the rating of a citizen for a given route, we subtract from each feature (distance, elevation and pavement quality) all the ``modifiers'' related to health
conditions and age (\textit{cf.} Table~\ref{tab:tableSkills}). Next, the resulting values of each feature are aggregated. Since each feature is in a rage
between 0 and 5, this results in a value between 0 and 15 (where 0 is the
worst value and 15 the best). Then, the result
is scaled to the widely used range $[0,10]$.
Also, to introduce some variability we add Gaussian noise sampled from
$\mathcal{N}(0,1.5)$.


\begin{table}[t]
\renewcommand{\arraystretch}{1.3}
\caption{Route features modifiers depending on health status and age.}
\label{tab:tableSkills}
\centering
\begin{tabular}{|>{\bfseries}c| c|c| c|}
\cline{2-4}
     \multicolumn{1}{c|}{}      &  \textbf{Distance} & \textbf{Elevation} &\textbf{Pavement} \\
\hline
18$<$Age$\leq$35 &          0 &          0 &          0 \\
\hline
35$<$Age$\le$50 &         -1 &         -1 &          0 \\
\hline
50$<$Age$\le$65 &         -2 &         -2 &          0 \\
\hline
65$<$Age &         -3 &         -3 &          -1 \\
\hline
Visual Impairment &          0 &          0 &          -1 \\
\hline
Breathing Problems &          0 &         -1 &          0 \\
\hline
Reduced Mobility &         -1 &         -1 &          -3 \\
\hline
Heart Diseases &         -1 &         -2 &          0 \\
\hline
\end{tabular}
\end{table}

\section{Conclusions and Future Work}
\label{sec:conclusions}

In this article we provide an overview of the internal details of our 
smart health application. Moreover, we detail the steps in order to generate a statistically sound simulated dataset with real medical data, in order to further validate the system. Therefore, a formal presentation and definition of the proposal, details about the main actors, real-time data processing and communication with the smart city, as well as validations, experiments and pilot tests are left to future work.

\balance

\bibliographystyle{IEEEtran}       
\bibliography{IEEEabrv,recommender_shealth}

\end{document}